\documentclass[9pt,conference]{IEEEtran}

\IEEEoverridecommandlockouts
\usepackage{cite}
\usepackage{amsmath,amssymb,amsfonts}
\usepackage{algorithmic}
\usepackage{graphicx}
\usepackage{textcomp}
\usepackage{xcolor}
\usepackage{todonotes}
\usepackage{booktabs}
\usepackage{graphicx}
\usepackage{booktabs}
\usepackage{adjustbox}
\usepackage{multirow}
\usepackage{url}
\usepackage{caption} 
\usepackage{etoolbox}

\def\BibTeX{{\rm B\kern-.05em{\sc i\kern-.025em b}\kern-.08em
    T\kern-.1667em\lower.7ex\hbox{E}\kern-.125emX}}

\begin{document}


\title{SENSE models: an open source solution for
 multilingual and multimodal semantic-based tasks 
}
\author{\IEEEauthorblockN{Salima Mdhaffar}
\IEEEauthorblockA{\textit{LIA} \\
\textit{Avignon Université}\\
Avignon, France \\
0000-0002-8472-6890}
\and
\IEEEauthorblockN{Haroun Elleuch}
\IEEEauthorblockA{\textit{Elyadata} \\
\textit{LIA/Avignon Université}\\
Tunis, Tunisia \\
0009-0006-1175-650X}
\and
\IEEEauthorblockN{Chaimae Chellaf}
\IEEEauthorblockA{\textit{Lundi Matin} \\
\textit{LIA/Avignon Université}\\
Avignon, France \\
0009-0001-7514-0740}
\and
\IEEEauthorblockN{Ha Nguyen$^*$\thanks{$^*$This work was performed while at LIA, Avignon Université} }
\IEEEauthorblockA{\textit{Oracle} \\
France \\
0009-0003-7813-1713}
\and
\IEEEauthorblockN{Yannick Estève}
\IEEEauthorblockA{\textit{LIA} \\
\textit{Avignon Université}\\
Avignon, France \\
0000-0002-3656-8883}
}

\maketitle

\begin{abstract}

This paper introduces SENSE (Shared Embedding for N-lingual Speech and tExt), an open-source solution inspired by the SAMU-XLSR framework and conceptually similar to Meta AI’s SONAR models. 
These approaches rely on a teacher–student framework to align a self-supervised speech encoder with the language-agnostic continuous representations of a text encoder at the utterance level. 
We describe how the original SAMU-XLSR method has been updated by selecting a stronger teacher text model and a better initial speech encoder. 
The source code for training and using SENSE models has been integrated into the SpeechBrain toolkit, and the first SENSE model we trained has been publicly released. 
We report experimental results on multilingual and multimodal semantic tasks, where our SENSE model achieves highly competitive performance. 
Finally, this study offers new insights into how semantics are captured in such semantically aligned speech encoders.

\end{abstract}


\section{Introduction}

Speech foundation models based on self-supervised learning (SSL) have brought significant advances in speech processing. 
These models, such as wav2vec 2.0 \cite{baevski2020wav2vec}, HuBERT \cite{hsu2021hubert}, and WavLM \cite{chen2022wavlm},  generate learned speech representations that can be applied to a wide range of downstream speech processing tasks.
By training on large amounts of unlabelled speech data, SSL models have demonstrated the ability to capture crucial speech features, such as phonemes and other acoustic units~\cite{pasad2023comparative}. 
This capability has led to significant progress in multiple downstream tasks, including speech recognition~\cite{baevski2020wav2vec}, speech translation~\cite{nguyen2020investigating}, speech separation, speaker verification, speaker diarization~\cite{chen2022wavlm}, and emotion detection~\cite{macary2021use}.

Different approaches have been proposed to pretrain model by aligning speech and text, like mSLAM~\cite{bapna2022mslam}, a \textsl{Massively multilingual joint pre-training for speech and text}.
In 2022, MIT and LIUM  introduced a novel approach to learn semantically-aligned, multimodal, utterance-level, cross-lingual speech representations (SAMU-XLSR: \textsl{Semantically-Aligned Multimodal Utterance-level cross-lingual speech representation})\cite{khurana2022samu} from SSL pretrained speech encoders. 
The embedding vector space of SAMU-XLSR is both multimodal and cross-lingual: it is shared between speech and text modalities and is common to multiple languages. 
Moreover, it is semantically aligned, as spoken utterances are clustered together with their corresponding speech and text translations within the vector space.
To achieve this alignment, \cite{khurana2022samu} proposes a multimodal learning framework that fine-tunes the pre-trained multilingual XLSR-128 speech encoder~\cite{babu2021xls} through a teacher-student approach led by the pre-trained language-agnostic BERT sentence encoder, LaBSE~\cite{feng2022language}. 
More recently, META AI has also presented the SONAR approach~\cite{Duquenne:2023:sonar_arxiv} as a part of the Seamless models for speech translation~\cite{seamless2025joint}.
The SAMU-XLSR and SONAR approaches are very similar since they both consist of making the models able to extract and represent semantics at the utterance level by learning from a textual model (respectively LaBSE~\cite{feng2022language} and a specialized NLLB decoder~\cite{nllbteam2022languageleftbehindscaling}).
They allow reaching very good results on tasks that require the extraction of semantics from speech (speech translation, spoken language understanding, speech retrieval\ldots).
While this approach is very promising, there is no open-source solution that could help researchers aiming to work in this direction. 
There is no public implementation of SAMU-XLSR, and even if the SONAR models are distributed under an open-source licence and the source code to use them, the code to retrain similar models is not available.


This paper introduces SENSE (Shared Embedding for N-lingual Speech and tExt), an open-source solution directly inspired by the SAMU-XLSR framework. 
We describe how the original SAMU-XLSR method has been updated by selecting a stronger teacher text model and a better initial speech encoder. 
The source code for training and using SENSE models has been integrated into the SpeechBrain toolkit~\cite{ravanelli2024open}, and the first SENSE model we trained has been publicly released\footnote{\url{https://huggingface.co/LIA-AvignonUniversity}}. 
For reproducibility reasons, we release the first SENSE model that have been trained using publically-available data.
We report experimental results on multilingual and multimodal semantic tasks, where our initial SENSE model achieves highly competitive performance. 
Finally, this study offers new insights into how semantics are captured in such multilingual, semantically aligned speech encoders.



\section{SAMU-XSLR framework}

\begin{figure}[h]
\centering
\includegraphics[width=0.65\columnwidth]{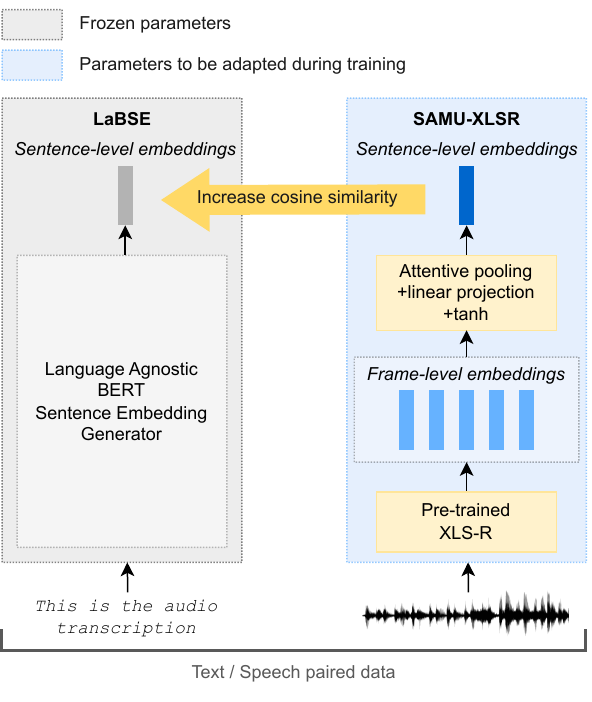}
\caption{SAMU-XLSR training framework}
\label{fig:trainingSAMUXLSR}
\end{figure}

SAMU-XLSR has been introduced~\cite{khurana2022samu} in order to fine-tune the multilingual XLS-R SSL speech encoder~\cite{babu2021xls}.
Frame-level representations are merged by an attentive pooling layer, then passed through a linear projection and \textsl{tanh} activation function, producing one utterance-level vector per audio file while keeping XLS-R weights adaptable (Fig.~\ref{fig:trainingSAMUXLSR}).

This speech vector is trained to imitate the LaBSE text embedding of the utterance transcript~\cite{feng2022language}.
The cosine similarity between the LaBSE text embedding and the utterance-level SAMU-XSLR is maximized through the loss function.
Because LaBSE supplies a language-agnostic space covering 109 languages, SAMU-XLSR learns cross-lingual speech–text alignment by simply pulling each speech vector toward its multilingual text anchor, with no explicit cross-lingual audio supervision.

\section{The first SENSE model}
\subsection{Teacher model selection}

The Massive Text Embedding Benchmark (MTEB) \cite{muennighoff2023mteb} aims to provide a comprehensive evaluation of text embedding models across a wide range of tasks.
It includes 58 datasets spanning 112 languages and covers 8 different embedding tasks.
We used the MTEB leaderboard\footnote{\url{https://huggingface.co/spaces/mteb/leaderboard}} to identify the model that best meets our specific requirements.
Among the top-performing models, BGE-M3 \cite{chen2024m3} stood out due to its strong performance in various tasks such as reranking, bitext mining, and semantic textual similarity.
In addition to its performance, BGE-M3 is relatively lightweight with 560 million parameters, fully open-source, and multilingual, supporting over 100 languages. 
It can also handle documents of varying lengths, with support for sequences of up to 8,192 tokens.

\subsection{SSL speech encoder selection}
The ML-SUPERB Challenge (Multilingual Speech Universal PERformance Benchmark) \cite{yang2024mlsuperb} provides a large-scale benchmark for evaluating multilingual speech processing systems, covering many languages.
Based on the 2025 baseline results \footnote{\url{https://multilingual.superbbenchmark.org/challenge-interspeech2025/challenge_overview}} trained using the ML-SUPERB 1-hour dataset, we observe that w2v-BERT 2.0 outperforms other models with fewer than 1 billion parameters.
A benchmark on Tunisian Arabic \cite{mdhaffar2024performance} has shown that the best open-source speech encoder is  w2v-BERT 2.0.
Internal benchmarks were conducted on different languages and across various acoustic conditions (clean and noisy data) and confirmed this finding.

\subsection{SENSE training details}
We developed our own implementation in SpeechBrain to build models of the SAMU-XLSR type.
As mentioned above, we selected a new teacher model, BGE-M3 \cite{chen2024m3}, and a new speech encoder model  w2v-BERT 2.0 \cite{barrault2023seamless} to initialize the SENSE model.
We then trained it on the publicly available Common Voice 19 dataset \cite{ardila2020common}, using only the languages supported by the BGE-M3 model, a total of 83 languages, corresponding to 8,250 hours of speech.
The Common Voice dataset is organized into two sets: validated and unvalidated.
The validated partition consists of audio samples that have been reviewed and confirmed by human annotators to have accurate transcriptions and good audio quality. 
Within the validated set, the data is further divided into train, development, and test subsets.
To train the SENSE model, we used only the validated training partition.
Details regarding the languages and the duration of each language subset are provided in the appendix \ref{app:lang}. 
Following \cite{khurana2022samu}, we perform either up- or down-sampling based on each language's representation in the dataset. 
We initialized our speech encoder model with the w2v-BERT 2.0 model. 
The w2v-BERT 2.0 model comprises 24 Conformer layers \cite{gulati2020conformer} and combines contrastive and masked prediction learning and is pre-trained on 4.5 million hours of unlabelled audio data that covers 143 languages.
In order to build our fixed-size utterance representation, we added an attention pooling \cite{santos2016attentive} operation on the encoder outputs.
This model is pre-trained for 100K iterations by employing 32 H100 GPUs. 
We select the batch size to optimize GPU memory usage, resulting in 20 min of audio per batch. 
We use different learning rates and optimizers for the pre-trained encoder and the attention pooling. 
We fine-tune the speech encoder using a learning rate of $10^{-5}$ while the attention pooling is trained with a learning rate of 1.
We used the cosine distance between the speech and the
text embedding as the training loss.
The text encoder is kept frozen throughout the training.



\section{Experimental Motivations}

Our primary aim is to assess how well the SENSE model can extract and represent semantic information from multilingual, multimodal sources. To place its capabilities in context, we benchmark SENSE against SAMU‐XSLR and SONAR speech encoders. 
Both of these models have demonstrated strong performance in aligning speech and text across multiple languages. 

\subsection{Cross‐lingual and Multimodal Retrieval}

To gauge the model’s ability to form truly language‐independent semantic representations, we evaluate its performance on multilingual, multimodal retrieval. 
In this setting, spoken utterances in one language must be matched to written translations in another language. 
By treating speech and text as two modalities within a shared embedding space, we test whether SENSE can bridge acoustic signals and textual content, regardless of language, solely based on learned semantic structure. 
Success on this task indicates that the model has abstracted away from language‐specific phonetic or syntactic details and is focusing instead on core meaning.

\subsection{Spoken Language Understanding and Summarization}

Beyond retrieval, it is crucial to verify that SENSE captures detailed semantic content from continuous speech. 
To this end, we employ Spoken Language Understanding (SLU) tasks that require extracting structured information such as intent, named entities, or slot values directly from audio. 
In parallel, we investigate the model’s capacity for higher‐level abstraction by evaluating speech summarization. 
Summarization requires condensing entire utterances into concise, coherent descriptions, testing whether SENSE can integrate information over longer time spans and identify salient content. 
Together, SLU and summarization reveal whether frame‐level and discourse‐level semantics are faithfully encoded.

\subsection{Speech‐to‐Text Translation}

Speech translation is a particularly relevant task to assess how well SENSE’s speech encoder captures cross‐lingual semantics. 
 By measuring translation quality (e.g., BLEU scores) on the resulting outputs, we evaluate whether SENSE’s encoder has distilled enough language‐independent meaning into its representations to support accurate translation. 
In other words, if the speech‐derived embeddings retain the essential semantics of the utterance, the downstream translator can produce fluent, correct text in the target language. 

By combining these evaluations (retrieval, SLU, summarization, and translation) and by comparing against SAMU‐XSLR and SONAR, we obtain a comprehensive picture of how effectively SENSE abstracts meaning from multilingual speech and aligns it with text across modalities.

\section{Datasets and experimental framework}
\subsection{\textbf{Multilingual and Multimodal Translation Retrieval task}}
To perform translation retrieval, we construct two databases (DB): a query database and a search database.
The query database contains speech utterances in language X, while the search database consists of either speech or text in language Y, depending on the scenario.
We consider many scenarios: speech $\to$ speech retrieval and speech $\to$ text retrieval.
The objective is to retrieve the correct translation in a different language.
Depending on the scenario, each audio segment or text sentence in the query database is transformed into an embedding.
If the modality is speech, a speech encoder is used to extract the embedding at the utterance level. 
If the modality is text, a text encoder is applied to extract the sentence embedding.
Similarly, the entries in the search database are processed in the same way, using the appropriate encoder based on their modality.
Once both sets of data (query database and search database) are turned into embeddings that can be compared and normalized via mean subtraction, translation retrieval is performed by computing the similarity between the query embeddings and all entries in the search database.
Following \cite{khurana2022samu}, we evaluate the  translation retrieval task using Recall@1.

\paragraph{\textbf{X Speech $\to$ EN Speech retrieval}} We use the public VoxPopuli dataset \cite{wang2021voxpopuli}, which provides a large-scale multilingual corpus suitable for speech-to-speech translation and other tasks. 
We follow \cite{khurana2022samu} and use the same languages.
This setup involves speech queries in one language and a search DB composed of their corresponding translations in English.
The query DB and the search DB are equal in size.
\paragraph{\textbf{EN Speech $\to$ Y Speech retrieval}}
We use the same set of languages as in the previous setup. 
Here, the speech queries are in English, and the search database consists of speech utterances in a target language.
\paragraph{\textbf{X Speech $\to$ Y Speech retrieval}}
We select a subset of random language pairs from the VoxPopuli \cite{wang2021voxpopuli} dataset used previously, where $X \neq Y$, $X \neq \text{EN}$, and $Y \neq \text{EN}$. 
\paragraph{\textbf{EN Speech $\to$ Y Text}}
We use the publicly available MUST-C \cite{di2019must} dataset.
MUST-C is a large-scale multilingual corpus designed specifically for speech-to-text translation tasks.
The speech query DB is in English and the search DB is in another language.
We follow \cite{khurana2022samu} to create the query and search databases.

\paragraph{\textbf{X Speech $\to$ EN Text}}
We use the public MTEDx dataset \cite{salesky2021multilingual}.
In order to evaluate the generalization ability of the models to unseen languages, we additionally used the FLEURS dataset \cite{conneau2023fleurs}.

\paragraph{\textbf{X Speech $\to$ Y Text}}
We use the FLEURS dataset \cite{conneau2023fleurs} and MTEDx dataset \cite{salesky2021multilingual} by selecting a subset of language pairs.

\subsection{\textbf{Spoken Language Understanding tasks}}
\paragraph{\textbf{Named Entity Recognition (NER)}}
We evaluate our model on MSNER \cite{meeus2024msner}, a multilingual NER benchmark based on subsets of the VoxPopuli dataset in four languages (Dutch, French, German, and Spanish).
The dataset is annotated using the 18 entity classes defined in OntoNotes \cite{weischedel11272ontonotes}, providing a structured and linguistically rich evaluation setting.
We use the official partitions of the dataset. 


\paragraph{\textbf{Slot Filling}}
For the slot filling task, we use SpeechMassive \cite{lee2024speech}
SpeechMassive is a dataset designed to evaluate SLU capabilities in both monolingual and cross-lingual settings. 
The annotation format in SpeechMassive follows the SLURP schema \cite{bastianelli2020slurp}.
In total, the dataset defines 55 distinct slot types.
We used also the SLURP dataset.
 Since for each task, we include some evaluation for low-resource languages not covered by the learned model, we evaluate SLU also using TARIC-SLU \cite{mdhaffar2024taric}.
 The TARIC-SLU dataset is a publicly available resource designed to advance SLU in Tunisian Arabic.
 It comprises over 2K dialogues.
 These dialogues are annotated with 62 semantic slot types.
 We used also MEDIA dataset \cite{laperriere2022spoken}.

The study in \cite{laperriere2023use} demonstrated the effectiveness of the semantically aware frame-level speech representations of SAMU-XLSR for a specific semantic extraction task. In line with this work, we adopt a frame-level approach in our evaluation.
We formulate the SLU (for both NER and Slot filling) task as a character level prediction where slots or named entities are delimited by tag specific special characters, as in~\cite{ghannay2018end, mdhaffar2022end}.
As input, the neural network receives an audio file and the output is a transcription enriched
with semantic labels.
In addition to the speech encoder model, we incorporate an extra layer with 1024 neurons and
LeakyReLU as the activation function, followed by a fully-connected layer and a final 
softmax layer where each dimension corresponding to a character. 
 The weights of these two additional
layers were randomly initialized, while the weights
of the speech encoder part for SSL models of the
neural architecture were initialized using the pretrained weights.
We evaluate the NER and slot filling tasks using Named Entity Error Rate (NEER) \cite{mdhaffar2022end}, COncept Error Rate (COER) and Concept Value Error Rate (CVER) \cite{laperriere2022spoken}.
COER and NEER are computed similarly to Word Error Rate by taking into account only the semantic (or entities) labels in the reference and hypothesis annotations.
The CVER
computation is identical, but the occurrences of concept/value pairs are taken into account instead of the concept alone.

\subsection
{\textbf{Speech summarization}}
For Speech summarization, we use DECODA dataset \cite{bechet2012decoda}.
DECODA is a French human-human spoken conversation dataset composed of spoken dialogues.
Each conversation is accompanied by a manual transcription and a short summary that captures the key points discussed in the conversation. 
Building upon the work of \cite{akani2024unified}, we employ the same version of the DECODA dataset. 
In this task, we use a large-scale pretrained sequence-to-sequence model called BARThez \cite{eddine2021barthez}, specifically designed for French.
We trained the BARThez model to generate summaries from speech embeddings.
The first training stage uses the French MLSUM dataset, as it is one of the largest summarization datasets available in French. 
This step is essential to help the model adapt to sentence embeddings as input. 
The second training stage specializes the model on task-specific dataset. 
We fine-tune the model on DECODA using speech utterance embeddings, to generate summaries directly from speech.
This task is evaluated using ROUGE-L \cite{lin2004rouge} and BertScore \cite{zhang2019bertscore}.

\subsection{\textbf{Speech-To-Text translation}}
We evaluated speech-to-text translation task using the CoVoST~2 dataset~\cite{wang2021covost}. 
CoVoST~2 is derived from Common Voice~\cite{ardila2020common} with additionnal translations.

We use the standard encoder-decoder architecture for our translation model.  
We initialize the encoder with a speech encoder and decoder using mBart \cite{liu2020mbart}.
A feed-forward network projection layer is used to connect the encoder and decoder, bridging the two modules. 
BLEU is used to evaluate the translation task. \\

For reproducibility purposes, the full list of hyperparameters et more details are provided in the Appendix \ref{app:hyperparams}.



\section{Experimental results}
This section presents the experimental results on all tasks described above.
We compare SENSE with two state-of-the-art models, SAMU-XLSR and SONAR, for experiments based on utterance-level embeddings. For frame-level experiments, the comparison is made with w2v-BERT 2.0 and SAMU-XLSR.
The SAMU-XLSR used in our work is the version trained on 53 languages, following the same training procedure as the 25-language version presented in \cite{khurana2022samu}.
SENSE and SAMU-XLSR use a single multilingual speech encoder shared across all languages. 
SONAR, on the other hand, uses 37 language-specific encoders.
In all our experiments, we use the corresponding speech encoder when the language is available; otherwise, we use the English speech encoder.

\subsection{Multilingual and Multimodal Translation Retrieval task}

Table \ref{tab:Speech_speecH_retr} presents Recall@1 (R@1) scores for speech-to-speech translation retrieval across multiple language pairs using three systems: SENSE, SAMU-XLSR, and SONAR, evaluated on the VoxPopuli dataset. 
The table is organized into three sections: X→EN, EN→Y, and X→Y retrieval. In all these scenarios, SENSE consistently achieves the highest performance.
Although ``hr'' is an unseen language for all models during training, the results demonstrate that SENSE generalizes well, maintaining high retrieval performance even in zero-shot conditions.
Despite the fact that SONAR English speech encoder was pretrained on data that includes VoxPopuli, SENSE still outperforms it, highlighting the robustness of SENSE in both seen and unseen language settings.

\captionsetup{skip=1pt}
\begin{table}[h!]
\centering
\captionsetup{justification=centering, font=small}
\caption{R@1 scores speech $\to$ speech translation retrieval for various language pairs (VoxPopuli)}
\setlength{\tabcolsep}{5.5pt} 
\renewcommand{\arraystretch}{0.92} 
\begin{tabular}{l@{\hskip 4pt}c@{\hskip 4pt}c@{\hskip 4pt}c@{\hskip 4pt}c@{\hskip 4pt}c@{\hskip 4pt}c@{\hskip 4pt}c@{\hskip 4pt}c}
\toprule
& \multicolumn{8}{c}{\textbf{X Speech $\to$ EN Speech retrieval}} \\
\toprule
\textbf{R@1} & \textbf{fr-en} & \textbf{pl-en} & \textbf{nl-en} & \textbf{es-en} & \textbf{hr-en} & \textbf{de-en} & \textbf{ro-en} & \textbf{cs-en} \\
\midrule
\textbf{SAMU-XLSR}          & 94.76 & 93.61 & 90.2  & 94.53 & 55.81 & 90.26 & 84.22 & 84.7 \\
\textbf{SONAR}                 & 91.91 & 95.79 & 95.16 & 95.3     & 52.16 & 94.18 & 94.45 & 95.62 \\
\textbf{SENSE}              & \textbf{96.55} & \textbf{96.46} & \textbf{95.75} & \textbf{96.48} & \textbf{96.5} & \textbf{94.71} & \textbf{96.83} & \textbf{96.7} \\
\bottomrule
& \multicolumn{8}{c}{\textbf{EN Speech $\to$ Y Speech retrieval}} \\
\bottomrule
\textbf{R@1} & \textbf{en-fr} & \textbf{en-pl} & \textbf{en-nl} & \textbf{en-es} & \textbf{en-hr} & \textbf{en-de} & \textbf{en-ro} & \textbf{en-cs} \\
\midrule
\textbf{SAMU-XLSR}      & 95.37 & 93.52 & 89.89 & 94.98 & 54.25 & 89.99 & 83.41 & 84.2  \\
\textbf{SONAR}            & 91.43 & 95.57 & 94.65 & 95.1     & 52.36 & 93.82 & 74.29 & 95.5  \\
\textbf{SENSE}          & \textbf{96.54} & \textbf{96.25} & \textbf{95.71} & \textbf{96.37} & \textbf{96.31} & \textbf{94.12} & \textbf{97.16} & \textbf{97.09} \\
\bottomrule
\textbf{Size DB}            & 49882 & 18768 & 10713 & 36320    & 7394 & 59116 & 16265 & 11181 \\
\bottomrule
& \multicolumn{8}{c}{\textbf{X Speech $\to$ Y Speech retrieval}} \\
\bottomrule
\textbf{R@1} & \textbf{fr-de} & \textbf{hr-cs} & \textbf{ro-fr} & \textbf{hu-da} & \textbf{de-fr} & \textbf{cs-hr} & \textbf{fr-ro} & \textbf{da-hu} \\
\midrule
\textbf{SAMU-XLSR}      & 94.86 & 38.19 & 85.19  & 4.54  & 95.06 &  38.91 & 83.81 &  5.16 \\
\textbf{SONAR}            & 91.39 & 52.93  &  92.01 &  3.32 & 92.73 & 53.65 & 92.15 & 4.31  \\
\textbf{SENSE}          & \textbf{95.2} & \textbf{94.75} & \textbf{96.55} & \textbf{92.69} & \textbf{95.36} & \textbf{94.69} & \textbf{96.90} & \textbf{92.63} \\
\midrule
\textbf{size DB}      & 55656 & 4973 & 15745 & 3528 & 55656 & 4973 & 15745 & 3528  \\
\midrule
\end{tabular}
\label{tab:Speech_speecH_retr}
\end{table}

Table \ref{tab:r1_results_must_c} reports Recall@1 (R@1) scores for English speech to target-language text translation retrieval on MUST-C.
Among the three systems compared (SENSE, SAMU-XLSR, and SONAR), SONAR consistently achieves the highest scores across all language pairs and shows clear improvements over the other models.
SONAR's strong performance can be partly attributed to the fact that it was pretrained on data overlapping with the MUST-C benchmark, giving it an advantage in this setting.
Moreover, although SONAR’s speech encoder was trained exclusively on English speech, it exhibits strong representational quality, likely contributing to its superior results despite its lack of multilingual acoustic exposure.

\begin{table}[ht]
\setlength{\tabcolsep}{1.25pt} 
\centering
\captionsetup{justification=centering, font=small}
\caption{R@1 scores EN Speech $\to$ Y Text Translation retrieval (MUST-C)}
\begin{tabular}{lcccccccccccc}
\toprule
\textbf{R@1} & en-es & en-fr & en-de & en-ro & en-it & en-cs & en-fa & en-ar & en-ru & en-zh \\
\midrule
\textbf{SAMU-XLSR} & 87.03 & 85.83 & 86.08 & 86.19 & 85.52 & 85.18 & 81.6 & 82.97 & 81.72 & 78.35 \\
\textbf{SONAR} & \textbf{90.63} & \textbf{90.07} & \textbf{90.33} & \textbf{90.78} & \textbf{90.58} & \textbf{88.97} & \textbf{84.88} & \textbf{86.91} & \textbf{86.57} & \textbf{82.04} \\
\textbf{SENSE} & 87.01 & 85.87 & 86.04 & 86.08 & 85.3 & 85.36 & 81.91 & 82.29 & 82.21 & 76.3 \\
\midrule
\# DB query & 4417 & 4643 & 4663 & 4525 & 4482 & 3927 & 3548 & 3669 & 4429 & 3265 \\
\# DB search & 270K & 280K & 234K & 240K & 258K & 132K & 185K & 215K & 270K & 188K \\

\bottomrule
\end{tabular}
\label{tab:r1_results_must_c}
\end{table}

Table~\ref{tab:r1_y_to_en_mtedx_fleurs} reports R@1 scores for X speech to English text translation retrieval on MTEDx and FLEURS datasets. 
SENSE achieves the best performance across most language pairs, including unseen ones, demonstrating strong generalization capabilities. This is particularly evident on FLEURS, where languages like lb, my, and bs were not seen by any of the models during training.

\begin{table}[ht]
\centering
\captionsetup{justification=centering, font=small}
\caption{R@1 scores for X Speech $\to$ EN text translation retrieval across MTEDx and FLEURS datasets}
\setlength{\tabcolsep}{1.1pt}
\begin{tabular}{lccccccccccc}
\toprule
\textbf{Model} 
& \multicolumn{5}{c}{\textbf{MTEDx}} 
& \multicolumn{5}{c}{\textbf{FLEURS}} \\
\cmidrule(lr){2-6} \cmidrule(lr){7-11}
& it-en & fr-en & es-en & pt-en & ru-en 
& ml-en & lb-en & uz-en & bs-en & my-en \\
\midrule
SAMU-XLSR & 89.59 & 85.86 & 81.63 & 78.03 & 76.87 
          & 26.75 & 50.38 & \textbf{55.63} & 48.05 & 8.65 \\
SONAR     & 89.01 & 82.45 & \textbf{85.92} & 85.05 & \textbf{84.76} 
          & 22.36 & 39.98 & 54.12 & 27.27 & 8.19 \\
SENSE     & \textbf{90.69} & \textbf{87.01} & 84.88 & \textbf{86.69} & 83.06 
          & \textbf{62.59} & \textbf{60.88} & 55.40 & \textbf{60.39} & \textbf{14.11} \\
\bottomrule
\# DB query & 1930 & 2095 & 1916 & 2033 & 2105 & 957 & 933 & 861 & 924 & 879 \\
\# DB search & 26.5K & 32K & 38K & 33K & 7K & 957 & 933 & 861 & 924 & 879 \\
\bottomrule

\end{tabular}
\label{tab:r1_y_to_en_mtedx_fleurs}
\end{table}

Table~\ref{tab:r1_combined_no_en} reports R@1 scores for speech-to-text translation retrieval from various source languages to non-English target languages, using the MTEDx and FLEURS datasets. 
Across both datasets, SENSE consistently achieves the highest scores, particularly excelling on low-resource FLEURS pairs such as ny-ces (27.00), sd-fr (59.74), and xh-ar (36.67).

\begin{table}[ht]
\centering
\captionsetup{justification=centering, font=small}
\caption{R@1 scores for Speech X $\to$ Text Y translation retrieval}
\setlength{\tabcolsep}{1.25pt}
\renewcommand{\arraystretch}{1.1}
\begin{tabular}{lccccccccccc}
\toprule
\textbf{Model} 
& \multicolumn{6}{c}{\textbf{MTEDx}} 
& \multicolumn{3}{c}{\textbf{FLEURS}} \\
\cmidrule(lr){2-7} \cmidrule(lr){8-10}
& it-es & es-fr & es-it & fr-pt & es-pt & pt-es 
& ny-ces & sd-fr & xh-ar \\
\midrule
SAMU-XLSR & 94.14 & 83.78 & 82.21 & 88.93 & 85.04 & 81.20 
          & 11.94  & 25.48 & 13.10 \\
SONAR     & 92.25 & \textbf{87.76} & \textbf{88.57} & 84.27 & 88.39 & 87.08 
          & 18.18  & 21.20 & 16.07 \\
SENSE     & \textbf{94.35} & 87.35 & 86.83 & \textbf{90.08} & \textbf{88.87} & \textbf{89.31} 
          & \textbf{27.00}  & \textbf{59.74} & \textbf{36.67} \\
\bottomrule
\# DB query & 1930 & 1917 & 283 & 2095 & 1917 & 2033 & 737 & 934 & 840 \\
\# DB search & 4190 & 5579 & 5882 & 15K & 23K & 13.5K & 737 & 934 & 840 \\
\bottomrule
\end{tabular}
\label{tab:r1_combined_no_en}
\end{table}

\vspace{-0.5cm}

\subsection{Spoken Language Understanding tasks}
\subsubsection{Named Entity Recognition}

\begin{table*}[ht]
\centering
\captionsetup{justification=centering, font=small}
\caption{Named Entity Error Rate (NEER) results on the MSNER dataset. Diagonal entries represent in-language training performance, while off-diagonal entries indicate zero-shot transfer performance. The best in-language NEER scores (diagonal) are underlined. The best zero-shot NEER scores per target language (columns) are shown in bold.}
\setlength{\tabcolsep}{3pt}
\renewcommand{\arraystretch}{0.9}
\begin{tabular}{lcccccccccccccccc}
\toprule
\multirow{1}{*}{}{} & \multicolumn{4}{c}{SAMU-XLSR} & \multicolumn{4}{c}{SONAR} & \multicolumn{4}{c}{w2v-BERT 2.0} & \multicolumn{4}{c}{SENSE} \\
\cmidrule(lr){2-5} \cmidrule(lr){6-9} \cmidrule(lr){10-13} \cmidrule{14-17}
        & NL   & ES    & FR    & DE    & NL   & ES    & FR    & DE    & NL   & ES    & FR    & DE    & NL   & ES    & FR    & DE \\
\midrule
NL      & \textit{22.80} & 100 & 100 & 99.61 & \underline{\textit{18.40}} & 93.85 & 98.56 & 86.50 & \textit{20.44} & 52.83 & 57.70 & 37.50 & \underline{\textit{18.40}} & \textbf{37.81} & \textbf{41.67} & \textbf{33.33} \\
ES      & 100 & \textit{27.84} & 100 & 100 & 90.31 & \textit{25.33} & 65.87 & 92.09 & 60.19 & \textit{25.61} & 41.45 & 47.77 & \textbf{36.85} & \underline{\textit{23.20}} & \textbf{35.94} & \textbf{36.90} \\
FR      & 100 & 100 & \textit{36.38} & 100 & 86.98 & 50.6 & \textit{35.17} & 85.42 & 63.31 & 50.84 & \textit{31.48} & 56.70 & \textbf{48.09} & \textbf{43.90} & \underline{\textit{30.27}} & \textbf{37.45} \\
DE      & 84.28 & 100 & 100 & \textit{32.00} & 33.4 & 92.94 & 98.4 & \textit{24.21} & 34.06 & 50.80 & 57.45 & \textit{23.24} & \textbf{28.82} & \textbf{35.13} & \textbf{42.55} & \underline{\textit{23.24}} \\
\bottomrule
\end{tabular}
\label{tab:ner_error}
\end{table*}

\begin{table*}[ht]
\centering
\setlength{\tabcolsep}{3pt} 
\captionsetup{justification=centering, font=small}
\caption{BLEU Speech X $\to$ English Text Translation results using CoVoST-2 dataset}
\setlength{\tabcolsep}{1.1pt}
\renewcommand{\arraystretch}{0.9}
\begin{tabular}{lccccccccccccccccccccc}
\toprule
& \textbf{FR} & \textbf{DE} & \textbf{ES} & \textbf{CA} & \textbf{IT} & \textbf{RU} & \textbf{Zh} & \textbf{PT} & \textbf{FA} & \textbf{ET} & \textbf{MN} & \textbf{NL} & \textbf{TR} & \textbf{AR} & \textbf{SV} & \textbf{LV} & \textbf{SL} & \textbf{TA} & \textbf{JA} & \textbf{ID} & \textbf{CY} \\
\midrule
\textbf{Duration (h)} & 180 &	119	 &97 &	81 &	28	& 16	& 10	& 7	 & 5 &	3&	3	&2	&2&	2&	2	&2&	2&	2	&1	&1	&1 \\
\midrule
\textbf{SAMU-XLSR} & 20.56 & 19.01 & 21.20 & 19.06 & 17.21 & 23.06  & 6.17  & 21.5 & 5.02 & 5.02 & 1.81 & 6.29 & 11.61 & 14.80 & 10.68 & 5.37 & 8.47 & 1.75 & 5.32 & 13.28 & \textbf{11.67} \\
\textbf{w2v-BERT 2.0} & 21.02 & 20.40 & 22.16 & 19.83 & 18.29 & 24.07 & 2.66 & 20.66 & 4.12 & 0.21 & 0.13 & 3.69 & 4.00 & 0.65 & 0.64 & 0.43 & 0.96 & 0.26 & 0.31 & 0.33 & 0.38 \\
\textbf{SENSE} & \textbf{21.67} & \textbf{21.95} & \textbf{23.24} & \textbf{21.57} & \textbf{20.69} & \textbf{27.96} & \textbf{6.33} & \textbf{26.16} & \textbf{7.49} & \textbf{9.27} & \textbf{2.25} & \textbf{15.00} & \textbf{14.36} & \textbf{15.82} & \textbf{16.81} & \textbf{8.35} & \textbf{12.11} & \textbf{2.54} & \textbf{6.22} & \textbf{14.12} & 8.65 \\
\bottomrule
\label{tab:bleu_s2t}
\end{tabular}
\end{table*}

Table~\ref{tab:ner_error} presents Named Entity Error Rates across four languages (NL, ES, FR, DE) using four different models: SENSE, w2v-BERT 2.0, SAMU-XLSR, and SONAR. 
The diagonal entries represent in-language training scenarios where the model was trained and evaluated on the same language, while off-diagonal entries correspond to zero-shot transfer, where the model was trained on one language and tested on another.
SENSE consistently shows the lowest error rates, both on the diagonal (e.g., NL $\to$ NL: 18.40\% NEER, ES $\to$ ES: 23.20\% NEER) and in zero-shot settings (e.g., ES $\to$ FR: 35.94\% NEER, DE $\to$ FR: 42.55\% NEER).
These results underline the superior cross-lingual transfer capabilities of SENSE, particularly in preserving semantic detection capabilities in multilingual NER.
In zero-shot settings (off-diagonal entries), SAMU-XLSR shows very poor generalization, with many error rates close to or at 100\%.
Error analysis revealed that a substantial proportion of the errors were substitution errors, where the system detected a slot but assigned the wrong label.

\subsubsection{\textbf{Slot Filling}}
Table \ref{tab:slu_finetune} reports the COER and CVER on four SLU datasets using full fine-tuning of four models. 
Across all settings, SENSE consistently yields the lowest error rates, outperforming w2v-BERT 2.0, SAMU-XLSR, and SONAR.

\begin{table}[h!]
\centering
\captionsetup{justification=centering, font=small}
\caption{COER and CVER (\%) for slot filling task}
\setlength{\tabcolsep}{3pt}
\renewcommand{\arraystretch}{0.85}
\begin{tabular}{lcccccc}
\toprule
\textbf{Model} & \textbf{Lang} & \textbf{Source} & \textbf{Size Train (h)} & \textbf{COER} & \textbf{CVER} \\
\midrule
\multirow{3}{*}{w2v-BERT 2.0} 
& FR & Speech Massive & 12.42 & 28.31 & 61.31 \\
& EN    & SLURP          & 13    & 15.84 & 42.39 \\
& TN    & TARIC          & 7.5    & 29.13 & 46.22 \\
& FR    & MEDIA          &  16.56   & 19.9 & 29.00 \\
\midrule
\multirow{3}{*}{SAMU-XLSR} 
& FR & Speech Massive & 12.42    & 32.77 & 66.74 \\
& EN    & SLURP          & 13    & 19.44 & 50.81 \\
& TN    & TARIC          & 7.5    & 30.11 & 48.06 \\
& FR    & MEDIA          &   16.56  & 18.7 & 29.4 \\
\midrule
\multirow{3}{*}{SONAR} 
& FR & Speech Massive & 12.42    & 35.86 & 74.59 \\
& EN    & SLURP          & 13    & 19.74 & 53.40 \\
& TN    & TARIC          & 7.5   & 33.58 & 52.33 \\
& FR    & MEDIA          & 16.56    & 20.2 &  31.74 \\
\midrule
\multirow{3}{*}{SENSE} 
& FR & Speech Massive   & 12.42    & \textbf{27.56} & \textbf{59.46} \\
& EN    & SLURP          & 13    & \textbf{14.00} & \textbf{38.98} \\
& TN    & TARIC          & 7.5    & \textbf{27.19} & \textbf{45.78} \\
& FR    & MEDIA          & 16.56    & \textbf{17.58} & \textbf{26.75} \\
\bottomrule
\end{tabular}
\label{tab:slu_finetune}
\end{table}

\subsection{Speech summarization}
Table \ref{tab:decoda_summ} presents the ROUGE and BERTScore results on the DECODA test-set. 
Notably, the best performance is achieved when the BARThez model is trained using SENSE speech embeddings for abstractive summarization. 
This configuration outperforms setups where speech embeddings are generated using SAMU-XLSR or SONAR.
Results show competitive results for SONAR.

\begin{table}[ht]
\centering
\captionsetup{justification=centering, font=small}
\caption{Evaluation results on test-set using ROUGE and BERTScore (F1) metrics.}
\begin{tabular}{lccc}
\hline
Model  & ROUGE-L & BERTScore-F1 \\
\hline
SAMU-XLSR  & 19.95 & 24.47 \\
SONAR  & 22.62 & 31.79 \\
SENSE  & \textbf{23.51} & \textbf{32.78} \\
\hline
\end{tabular}
\label{tab:decoda_summ}
\end{table}

\subsection{Speech-To-Text translation}

Table~\ref{tab:bleu_s2t} presents BLEU scores for speech-to-text translation across 21 target languages. 
The results compare three models: SENSE, SAMU-XLSR, and w2v-BERT 2.0. 
In this paper, we chose to evaluate translation using experiments based on the frame-level. 
We follow the approach of the SAMU-XLSR framework, which has shown that strong results can also be achieved using frame-level embeddings.
We did not include SONAR in our experiments, as its architecture is adapted to an NLLB decoder, unlike the other models.
Overall, SENSE outperforms the baselines on the majority of languages, particularly in low-resource cases such as FA, ET, MN, TA, and JA.
While all models perform reasonably well on high-resource languages (e.g., FR, DE, ES), w2v-BERT 2.0 shows strong performance only in a few mid-resource settings, but its scores collapse for many lower-resource targets (e.g., near-zero scores for ET, MN, and TA). 
SAMU-XLSR generally ranks between the two, showing better consistency than w2v-BERT 2.0 lower-resource targets. 
These results highlight SENSE’s robustness across both high- and low-resource languages, making it more suitable for generalized speech-to-text translation tasks.

\vspace{-0.2cm}

\section{Frame‐Level Semantic Attention Analysis}

In this section, we investigate how the general semantics of an utterance are captured at the frame‐level by the SENSE model. In particular, we check whether semantic information is uniformly distributed among all frame‐level embeddings or if specific patterns arise. The semantic information we study is that which allows the SENSE model to generate a sentence‐level embedding in the continuous space of the teacher text model. We assume that analyzing the attention logits used to weight the frame‐level embeddings when producing the sentence‐level embedding is a relevant indicator of where semantics are captured.

\begin{figure}[h!]
  \centering
  \includegraphics[width=0.8\columnwidth]{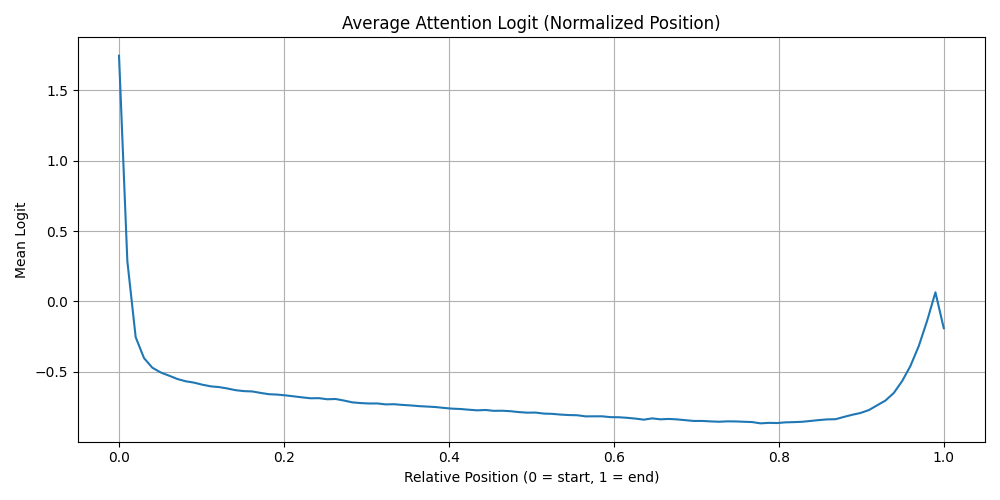}
  \caption{Average Attention Logit (Normalized Position)}
  \label{fig:att_avg_pos}
\end{figure}

Figure~\ref{fig:att_avg_pos} illustrates the general trend of attention logit distribution along the time axis. This figure shows the average logit value as a function of position in the sequence of frame embeddings. We generated this figure by extracting the attention logits from the 4096 French utterances in the 2009 Vox Populi corpus that have English speech translations. Since these utterances have different lengths, we normalized positions by resampling the sequence of attention values according to their relative position
$\textsl{relpos}(i) \;=\; \frac{i}{T-1}$, 
where $i$ is the absolute frame index (starting at $0$) and $T$ is the length of the sequence. The resampling is based on linear interpolation.

We observe that a significant portion of semantics is usually captured in the first few SENSE frame embeddings of an utterance, and, to a lesser extent, at the end of the sequence. The rest of the curve is relatively flat, with a slight decrease; however, since the curve represents means over thousands of utterances, it masks local variations arising from different grammatical patterns or words.

\begin{figure}[h!]
  \centering
  \includegraphics[width=0.9\columnwidth]{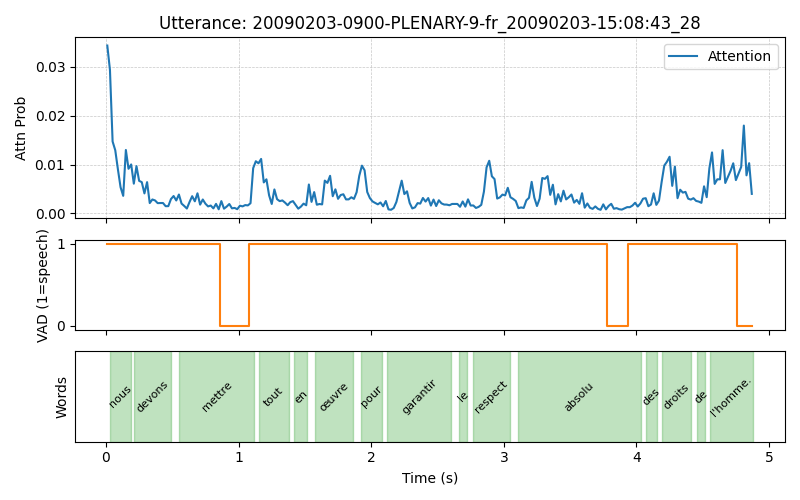}
  \caption{Example of attention‐weight distribution with VAD information and forced word alignment in French.}
  \label{fig:word+vad+attention_NoSilence}
\end{figure}

Figure~\ref{fig:word+vad+attention_NoSilence} shows the curve of attention weights applied to each frame embedding computed by the SENSE model for a single utterance. 
We can see the high peak at the beginning, as well as additional peaks appearing in regions where Figure~\ref{fig:att_avg_pos} was relatively flat. 
Additionally, the VAD curve (in orange) indicates the presence of non‐speech segments.
We continue our investigation by applying a voice activity detector (VAD) to the audio files. 
For this purpose, we relied on the Silero VAD tool~\cite{SileroVAD}. 
We also applied WhisperX~\cite{bain2022whisperx} to obtain forced word alignments. 
Although these word alignments are not very precise, they provide useful information for understanding how semantics are captured locally.

\vspace{-0.7cm}
\begin{figure}[h!]
  \centering
  \includegraphics[width=0.9\columnwidth]{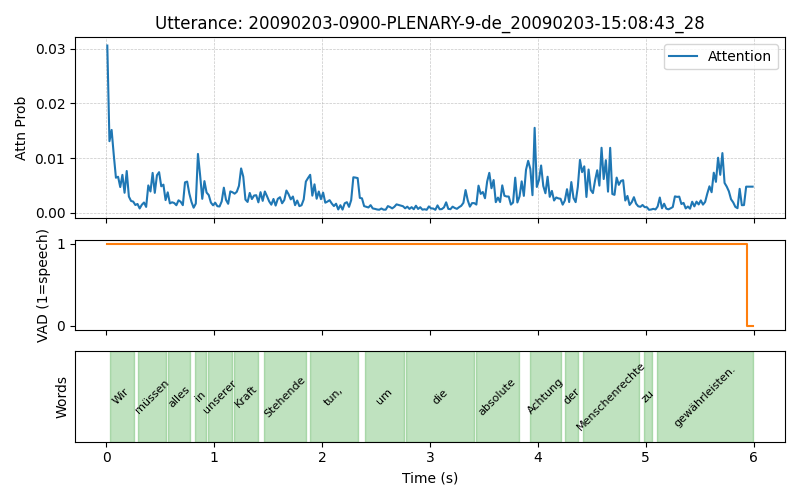}
  \caption{Example of attention‐weight distribution with VAD information and forced word alignment in German (spoken translation from French)}
  \label{fig:word+vad+attention_NoSilence_DE}
\end{figure} 
To quantify the importance of the first frames, we measured the average attention mass associated with the first five frame‐level embeddings. These first five embeddings concentrate 5.25\% of the attention weights, while they represent only 0.61\% of the average number of frames in an utterance.

The visualization of the word alignment helps explain the peaks observed in the attention curve: they usually correspond to the beginning of content words with high semantic information, while function words (determiners, some prepositions, etc.) do not trigger such behavior.
To further investigate multilingual aspects beyond French, we also examine the German utterances derived from the French ones in the VoxPopuli corpus. 
We observe similar behavior of the SENSE model across both languages, as illustrated in Figure~\ref{fig:word+vad+attention_NoSilence_DE}, with peaks of attention at the beginning of the utterances, and at the beginning of some real words.
The utterance in Figure~\ref{fig:word+vad+attention_NoSilence_DE} is the German spoken translation of the French utterance in Figure~\ref{fig:word+vad+attention_NoSilence}.

\begin{table}[h!]
\captionsetup{justification=centering, font=small}
\caption{Top 10 word occurrences by single‐utterance attention‐logit sum, and average attention‐logit sum for the 10 most frequent words.}
  \centering
  \begin{tabular}{l r  l r}
    \multicolumn{2}{c}{\textbf{Top 10 Occurrences}} &
    \multicolumn{2}{c}{\textbf{Avg Logit Sum }} \\
    \multicolumn{2}{c}{\textbf{ by Logit Sum}} &
    \multicolumn{2}{c}{\textbf{of 10 Most Frequent Words}} \\
    \hline
    \textbf{Word}           & \textbf{Sum}     & \textbf{Word}  & \textbf{Avg Sum}   \\
    \hline
    immunité             & 27.832413        & de            & -5.085950         \\
    Internet               & 23.647682        & la            & -4.170271         \\
    Mayotte               & 20.187487        & et            & -6.696034         \\
    Schengen               & 16.527946        & les           & -5.931593         \\
    Kosovo                 & 16.252068        & des           & -6.380726         \\
    radiodiffuseurs       & 15.779675        & le            & -4.103326         \\
    vidéoconférence        & 15.697246        & à             & -2.212737         \\
    Lisbonne             & 15.547420        & que           & -11.068631        \\
    Pagano              & 15.201697        & en            & -5.128108         \\
    Berlusconi            & 14.948489        & pour          & -9.791382         \\
    \hline
  \end{tabular}
  \label{tab:word_stats_att_logit}
\end{table}

Table~\ref{tab:word_stats_att_logit} shows the ten words observed with the highest sum of attention logits in a single utterance, alongside their respective summed values.
In the same table, we also report the average attention‐logit sum for the ten most frequent words across the corpus. 
Notably, the words with the highest single‐utterance sums, such as \emph{immunité} (immunity), \emph{Internet}, and \emph{Mayotte}, are relatively rare proper nouns or specialized terms. 
When these tokens appear, the model concentrates a disproportionate amount of attention on them, reflecting their strong, discriminative semantic content within those utterances. 
By contrast, highly frequent words like \emph{de} (of), \emph{la} (the), and \emph{et} (and) exhibit negative average logit sums, indicating that the model assigns relatively low logits to these function words. 
These common tokens do not carry significant semantic weight in the SENSE attention mechanism.

\section{Conclusion}

In this work, we introduced SENSE, an open-source framework designed to facilitate the development of multilingual and multimodal semantic-based models. By integrating our implementation into the SpeechBrain toolkit and publicly releasing our first trained model, we have made it possible for researchers to construct highly competitive state-of-the-art models for a variety of tasks, including multilingual retrieval, spoken language understanding, speech summarization, and speech-to-text translation.

Moreover, this paper presents, for the first time, a detailed investigation into how utterance-level semantics are derived from frame-level embeddings in such semantically aligned models. Our analyses highlight that semantic information is not uniformly distributed across embeddings but is often concentrated at specific points within speech utterances, emphasizing discriminative semantic content such as proper nouns and specialized terms.

Nevertheless, numerous questions remain open regarding the precise mechanisms behind semantic extraction in multilingual contexts and under varying acoustic conditions. We believe that the open-source nature of the SENSE framework presents an opportunity for the research community to further explore and deepen our collective understanding of these semantic embedding dynamics.

\section*{Acknowledgment}

This work was partially funded by the French Research Agency (ANR) through the PANTAGRUEL (ANR 23-IAS1-0001) project, the Institut Carnot Cognition through the ERSO project and ESPERANTO (grant agreement N°101007666).
It used HPC resources from GENCI-IDRIS: grants AD011012551R3,  AD011012108R3, AD011015509, A0171013801 and A0171014633.
The authors thank Antoine Laurent and Sameer Khurana for sharing one of their SAMU-XSLR models, and for the fruitful discussions.


%



\bibliographystyle{IEEEtran}
\bibliography{bibliographie}

\clearpage

\appendix


\subsection{List of languages and duration used to train SENSE}
\label{app:lang}
This section presents the list of languages used to train the SENSE model, along with their corresponding ISO codes and total durations of available speech data. The dataset covers a diverse range of languages with varying amounts of audio, as summarized in Table~\ref{tab:speech_durations_compact}.

\begin{table}[htbp]
\centering
\caption{Training Data Duration by Language}
\scriptsize
\begin{tabular}{|l|l|r||l|l|r|}
\hline
\textbf{Lang} & \textbf{ISO} & \textbf{Train} & 
\textbf{Lang} & \textbf{ISO} & \textbf{Train} \\
\hline
af & afr & 00:11:08 & am & amh & 00:48:24 \\
ar & arb & 32:21:50 & as & asm & 01:05:47 \\
ast &  & 00:30:18 & ba & bashk & 138:28:27 \\
be & bel & 472:35:42 & bg & bul & 07:05:32 \\
bn & ben & 32:46:34 & br & bre & 02:28:24 \\
ca & cat & 1748:12:47 & ckb &  & 08:59:10 \\
cs & ces & 27:13:10 & cv & chv & 01:59:10 \\
cy & cym & 11:26:32 & da & dan & 04:07:44 \\
de & deu & 928:06:24 & dv & div & 03:46:30 \\
el & ell & 02:07:56 & en & eng & 1743:18:52 \\
eo & epo & 235:04:29 & es & spa & 496:25:19 \\
et & est & 05:42:40 & fa & fas & 29:53:59 \\
fi & fin & 02:18:01 & fr & fra & 814:54:55 \\
fy-NL & fry & 05:31:05 & ga-IE & gle & 00:35:53 \\
gl & glg & 42:00:53 & gn & grn & 01:20:21 \\
he & heb & 01:22:10 & hi & hin & 05:47:27 \\
hsb &  & 01:27:17 & hu & hun & 53:41:56 \\
ia & ina & 04:30:53 & id & ind & 07:38:05 \\
it & ita & 251:08:54 & ja & jpn & 13:03:50 \\
ka & kat & 86:44:37 & kab &  & 143:29:23 \\
kk & kaz & 00:44:12 & ko & kor & 00:50:02 \\
ky & kir & 02:18:33 & lt & lit & 10:06:25 \\
lv & lav & 21:14:15 & ml & mal & 01:21:47 \\
mn & khk & 03:06:08 & mhr &  & 209:15:15 \\
mk & mkd & 02:05:56 & mr & mar & 01:49:35 \\
mt & mlt & 02:23:04 & ne-NP & npi & 00:21:48 \\
nl & nld & 45:29:51 & nn-NO & nno & 00:35:44 \\
oc & oci & 00:22:34 & or & ory & 02:40:10 \\
os & oss & 00:17:07 & pa-IN & pan & 01:09:32 \\
pl & pol & 29:47:25 & ps & pus & 02:23:12 \\
pt & por & 24:35:58 & ro & ron & 05:40:03 \\
ru & rus & 37:35:58 & sah &  & 03:04:22 \\
sc & srd & 00:44:20 & sk & slk & 03:31:20 \\
sl & slv & 01:20:46 & sr & srp & 00:36:42 \\
sv-SE & swe & 08:47:01 & sw & swh & 69:18:50 \\
ta & tam & 68:08:48 & th & tha & 37:09:31 \\
ti & tir & 00:01:16 & tk & tuk & 01:03:33 \\
tr & tur & 40:04:47 & tt & tat & 09:14:22 \\
ug & uig & 80:12:36 & uk & ukr & 30:22:51 \\
ur & urd & 08:07:29 & uz & uzn & 52:31:28 \\
vi & vie & 02:49:31 & yo & yor & 02:05:47 \\
zh-HK & zho & 09:29:54 &  &  &  \\
\hline
\end{tabular}
\label{tab:speech_durations_compact}
\end{table}

\subsection{Details of hyperparams for downstream tasks}
\label{app:hyperparams}
\subsubsection{Slot Filling and Named Entity Recognition}
For the fine-tuning phase, we adopt as a downstream probe a straightforward architecture consisting of a one layer DNN followed by a softmax activation function. 
The training is performed using the Connectionist Temporal Classification (CTC) loss. 
The probe's hidden layers have a dimension of 1024, with a dropout rate of 0.15.
We use different learning rates and optimizers for the pre-trained encoder and the probe. 
The speech encoder is fine-tuned using a learning rate of $1\times10^{-5}$ with a probe learning rate of $1.5$. 
The batch size is 4.
We fine-tune the entire models for 50 epochs.

\subsubsection{Speech summarization}
For the task of speech summarization, we trained the BARThez model to generate summaries from speech embeddings.
However, the BARThez model expects input sequences as vectors of size 768, whereas the embedding models we integrate may produce vectors of different sizes. 
To align these representations with BARThez's expected input space, we introduce a linear projection layer followed by a GeLU activation function, except for SAMU-XLSR embeddings, which already match BARThez's dimensionality and thus require only the activation function.
The first training stage uses the French MLSUM dataset, as it is one of the biggest summarization datasets available in French. 
This step is essential for helping the model adapt to sentence embeddings as input. 
We trained our model using the AdamW optimizer, with a batchsize of 16 and different learning rates for the pretrained seq2seq model and the projection layer. 
The pretrained seq2seq model parameters were optimized with a learning rate of $1\times10^{-5}$, while the linear projection layer was trained with a larger learning rate of $1\times10^{-3}$, allowing faster adaptation of the randomly initialized weights. 
A weight decay of $1\times10^{-5}$ was applied uniformly to regularize both components. We used the standard cross-entropy loss to train the model on this sequence prediction task. \\
The second training stage specializes the model on task-specific dataset. 
Thus, for speech summarization, we continue the training of the model on DECODA and evaluated the performance by using speech utterance embeddings generated from the test-set.
As the goal is to compare SONAR, SAMU-XLSR and SENSE embeddings in this task, we did the following trainings :
\begin{itemize}
    \item We trained the BARThez model on the MLSUM dataset using text sentence embeddings generated by the SONAR text encoder, and subsequently fine-tuned it on the DECODA dataset using SONAR speech embeddings.
    \item We trained the BARThez model on the MLSUM dataset using text sentence embeddings from the LaBSE text encoder, and then fine-tuned it on DECODA using SAMU-XLSR speech embeddings.
    \item We trained the BARThez model on the MLSUM dataset using text sentence embeddings from the BGE text encoder, followed by fine-tuning on DECODA using SENSE speech embeddings.
\end{itemize}

 \subsubsection{Speech Translation}
 We initialize the encoder with a speech encoder and decoder using mBart \cite{liu2020mbart}.
 A feed-forward network projection layer is used to connect the encoder and decoder, bridging the two modules. 
The speech encoder is fine-tuned with a learning rate of $1\times10^{-4}$, and the mBART decoder with $1\times10^{-4}$, while the feed-forward projection layer before the decoder is trained with a higher learning rate of $1\times10^{-3}$. All modules are optimized using the Adam optimizer.
We train the model over 100 epochs with a batch size of 4.

\end{document}